%% file: main.tex
\documentclass{article} 
\usepackage{iclr2024_conference,times}

\input{math_commands.tex}

\usepackage{amssymb}
\usepackage{hyperref}
\usepackage{url}
\usepackage{array}
\usepackage{graphicx}
\usepackage{booktabs}
\usepackage{arydshln}
\usepackage{multirow}
\usepackage{hyperref}
\usepackage{xcolor}

\usepackage{hyperref}
\newcommand{\source}[1]{\textcolor{cyan}{#1}}
\definecolor{darkgreen}{rgb}{0, 0.5, 0.0}
\newcommand{\reference}[1]{\textcolor{orange}{#1}}
\newcommand{\target}[1]{\textcolor{darkgreen}{#1}}
\newcommand{\imstart}{$ <\text{\textbar{im\_start}\textbar}>$}
\newcommand{\imend}{$<\text{\textbar{im\_end}\textbar}>$}
\title{SCALE: Synergized Collaboration of Asymmetric Language Translation Engines}
\author{Xin Cheng$^{1}$ \quad   Xun Wang$^{2}$ \quad Tao Ge$^{2}$ \\
\bf Si-Qing Chen$^{2}$\quad  Furu Wei$^{2}$ \quad Dongyan Zhao$^{1}$ \quad Rui Yan$^{3}$\\
\\
$^1$\ Peking University \quad $^2$\ Microsoft \quad  $^3$\ Renmin University of China }

\iclrfinalcopy 
\begin{document}
\maketitle

\begin{abstract}
In this paper, we introduce SCALE, a collaborative framework that connects compact Specialized Translation Models (STMs) and general-purpose Large Language Models (LLMs) as one unified translation engine. By introducing translation from STM into the triplet in-context demonstrations, SCALE unlocks refinement and pivoting ability of LLM, thus mitigating language bias of LLM and parallel data bias of STM, enhancing LLM speciality without sacrificing generality, and facilitating continual learning without expensive LLM fine-tuning.
Our comprehensive experiments show that SCALE significantly outperforms both few-shot LLMs (GPT-4) and specialized models (NLLB) in challenging low-resource settings. Moreover, in Xhosa to English translation, SCALE experiences consistent improvement by a 4 BLEURT score without tuning LLM and surpasses few-shot GPT-4 by 2.5 COMET score and 3.8 BLEURT score when equipped with a compact model consisting of merely 600M parameters. SCALE could also effectively exploit the existing language bias of LLMs by using an English-centric STM as a pivot for translation between any language pairs, outperforming few-shot GPT-4 by an average of 6 COMET points across eight translation directions. Furthermore we provide an in-depth analysis of SCALE's robustness, translation characteristics, and latency costs, providing solid foundation for future studies exploring the potential synergy between LLMs and more specialized, task-specific models\footnote{Code available at:  \url{https://github.com/Hannibal046/SCALE}}.
\end{abstract}

\section{Introduction}
Large Language Models (LLMs) have recently revolutionized the field of natural language processing \citep{OpenAI2023GPT4TR,touvron2023llama,peng2023rwkv}, significantly influencing machine translation (MT) by delivering exceptional performance without requiring a bilingual corpus, particularly in high-resource languages \citep{DBLP:journals/corr/abs-2005-14165,DBLP:journals/corr/abs-2302-01398}.
Moreover, as a unified multi-task learner, LLMs represent a substantial step towards artificial general intelligence \citep{bubeck2023sparks}, with the potential to overcome not only language barriers but also cultural boundaries simultaneously through a simple ``translate and explain" prompt.

Despite their advancements, LLM-based translation systems still confront several challenges. 
Firstly, there exists a significant language bias towards English (e.g., 92.1\% of the GPT-3 pre-training corpus is English, while French, the second largest, represents only 1.8\%\footnote{\url{https://github.com/openai/gpt-3/blob/master/dataset_statistics/languages_by_character_count.csv}}), which significantly constraints multilingual translation performance, especially for those low-resource languages \citep{scao2022bloom,DBLP:journals/corr/abs-2302-09210}.
Secondly, as a practical approach for system improvement, fine-tuning LLM poses great challenges. These include (1) the trade-off between speciality and generality \citep{cheng-etal-2023-decouple,lin2023speciality}, and (2) the prohibitively high cost associated with tuning large-scale models \citep{hu2021lora,dettmers2023qlora}.
In contrast, traditional Specialized Translation Models (STMs)---those based on encoder-decoder architecture, trained with supervision and significantly smaller in size \citep{sutskever2014sequence,vaswani2017attention}---serve as specialists for specific translation tasks and could be efficiently fine-tuned. However, these models lack general language capabilities and are potentially susceptible to parallel data bias, such as the memorization of low-quality samples \citep{raunak-etal-2022-salted}.

In this paper, we demonstrate for the first time the possibility to unify these two asymmetric translation engines in a single framework. Our work, SCALE, connects LLMs and STMS by utilizing the LLM's most enigmatic capability: in-context learning. Rather than employing source-target pairs as in conventional few-shot translation \citep{DBLP:journals/corr/abs-2302-01398,vilar2023prompting}, SCALE would first sample translations from a STM and then use triplets consisting of a source sentence, an STM-generated set and a target sentence as in-context demonstrations to unlock the refinement and pivoting ability of LLMs. With SCALE, we could
(1) mitigate both language bias of LLMs by utilizing an STM that concentrates on a specific language pair, and parallel data bias of STMs by using a general-purpose LLM as the main body of the system;
(2) enhance the speciality of LLMs without compromising generality;
(3) facilitate continual learning within the framework by updating only the lightweight STM, thus avoiding expensive LLM fine-tuning. 
By employing SCALE, we create a more efficient and effective system that combines the best of both translation engines.

\begin{figure*}[t!]
  \centering \includegraphics[width=0.9\textwidth,height=0.42\textwidth]{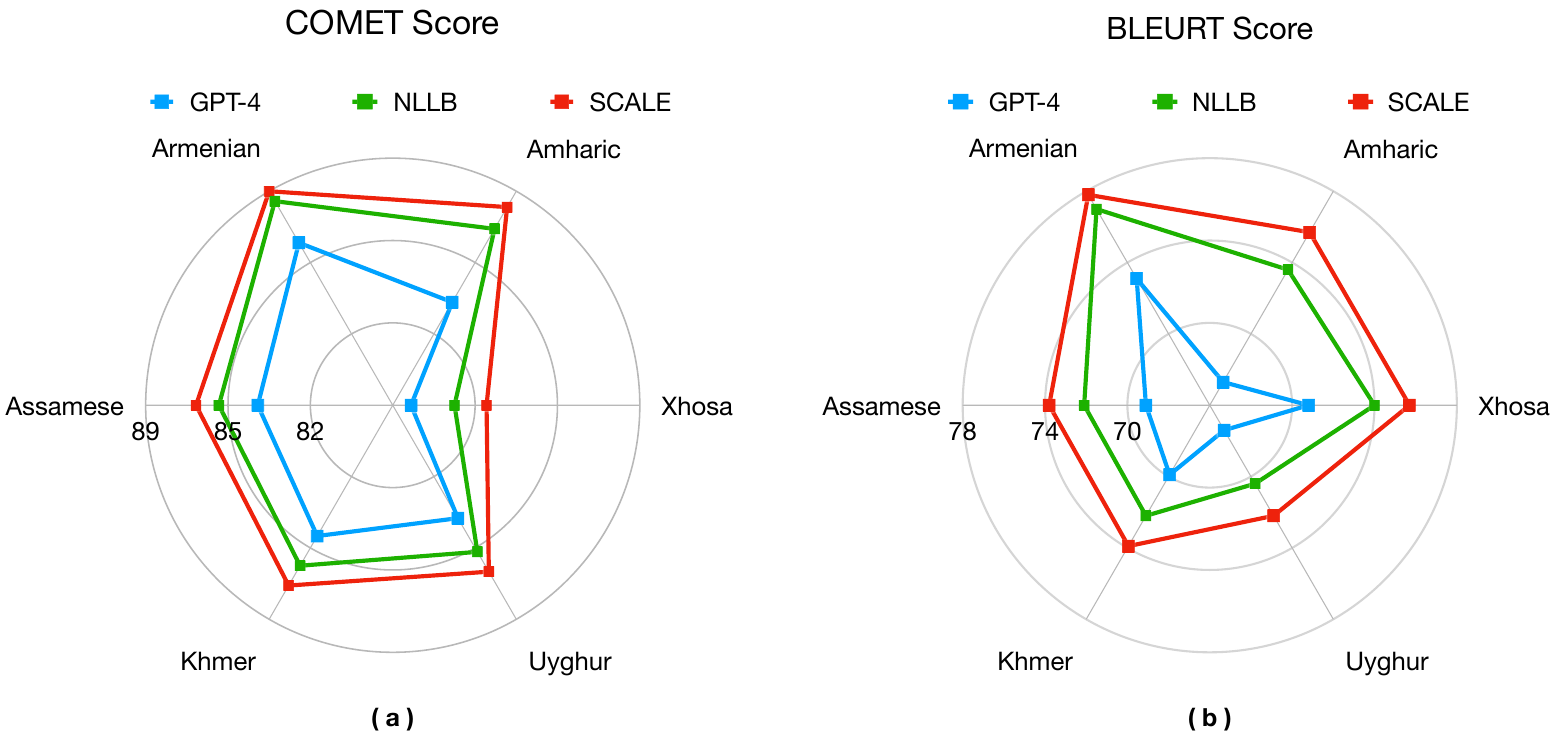}
  \caption{Translation results of few-shot LLM (GPT-4), STM (NLLB) and SCALE~(ours) for six low-resource languages measured by COMET and BLEURT.}
  \label{figure:cover}
\end{figure*}

Our comprehensive experiments reveal that SCALE considerably outperforms few-shot LLMs (e.g., GPT-4) and specialized models (e.g., NLLB) in the challenging low-resource setting, as depicted in Figure \ref{figure:cover}. Moreover, in Xhosa to English translation, SCALE experiences consistent improvement by a 4 BLEURT score without tuning LLM and surpasses few-shot GPT-4 by 2.5 COMET score and 3.8 BLEURT score when equipped with a compact model consisting of merely 600M parameters. Remarkably, SCALE can effectively exploit the existing language bias of LLMs by using an English-centric STM as a pivot for translation between any language pairs, outperforming few-shot GPT-4 by an average of 6 COMET points across eight translation directions.
Furthermore, we conduct an in-depth analysis of the robustness, translation characteristics, and latency costs associated with SCALE. Our findings provide valuable insights and encourage further research in this field.
 
\section{The SCALE Framework}
\begin{figure*}[thb!]
  \centering
  \includegraphics[width=0.9\textwidth,height=0.256\textwidth]{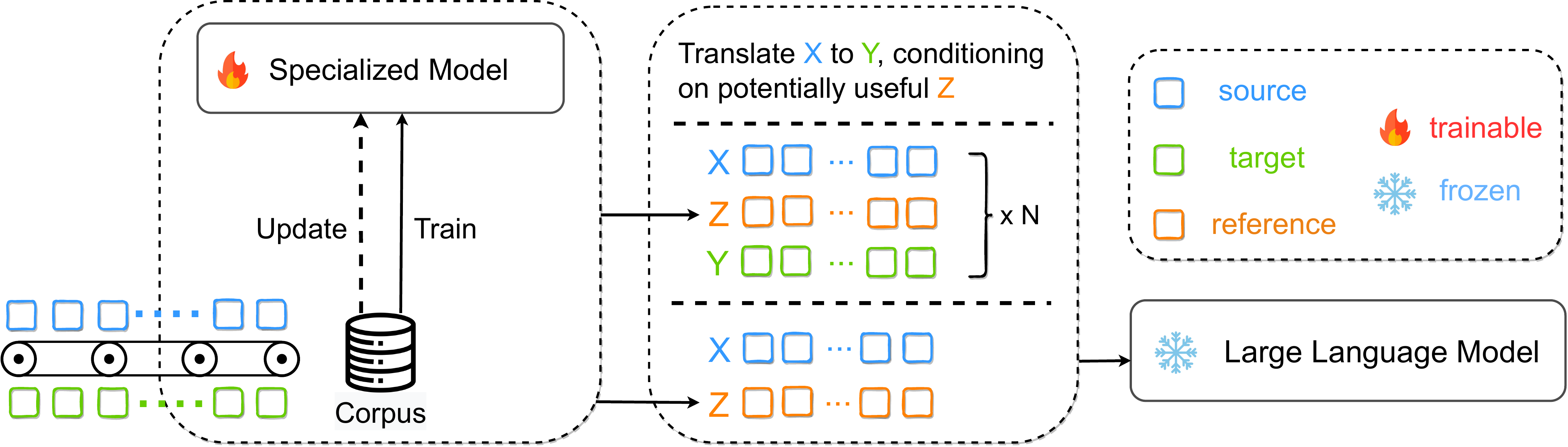}
  \caption{The SCALE framework, comprised of a lightweight specialized model and a frozen large language model with triplet in-context demonstrations.}
  \label{figure:framework}
\end{figure*}
In this section, we present the proposed SCALE method and provide an overview illustrated in Figure \ref{figure:framework}. Popularized by GPT-3~\citep{DBLP:journals/corr/abs-2005-14165}, In-context Learning~(ICL) allows LLMs to perform a wide variety of tasks, even newly created ones \citep{bills2023language}, by leveraging few-shot learning with a limited number of demonstrations. For a translation task from a source language $\mathcal{X}$ to a target language $\mathcal{Y}$, an LLM with parameters $\theta$ carries out ICL by conditioning on $k$ source-target paired examples $\mathbb{E} = (x_1,y_1)\oplus(x_2,y_2)\oplus ...(x_k,y_k)$ and the test source sentence $x$, generating the target $y$ in an auto-regressive manner as $y_{t} \sim p_{\theta}(y_{t}|\mathbb{E},x,y_{<t})$. In this scenario, the LLM must analyze the provided examples to discern the input distribution, output distribution, input-output mapping, and formatting to successfully complete the task \citep{press2022measuring,wei2023larger}. Different from conventional ICL, SCALE introduces an intermediate variable $\mathbb{Z}$ as reference between source $x$ and target $y$, transforming each demonstration example into a triplet $(x,\mathbb{Z},y)$. The variable $\mathbb{Z}$ is a generation set sampled from a specialized translation model $\mathbf{M_{\mathcal{X}\mapsto\mathcal{Y}}}$ trained on a labeled dataset. The final input to the LLM consists of the instruction, demonstrations, and source sentence combined in a prompt template: $
\mathcal{T}((x_1,\mathbb{Z}_1,y_1)\oplus(x_2,\mathbb{Z}_2,y_2)...\oplus(x_k,\mathbb{Z}_k,y_k)),(x,\mathbb{Z}))$. 
Unlike language understanding tasks that have fixed label set \citep{xu2023small}, the hypothesis space of translation model is actually infinite, so we could sample multiple generation paths from STM for one single source sentence to provide a more comprehensive generation guide for LLM.
The SCALE framework, though conceptually straightforward, demonstrates several advantages over STMs and LLMs, as highlighted below:

\textbf{Refinement} For $\mathcal{X}$ to $\mathcal{Y}$ translation task, when the intermediate variable $\mathbb{Z}$ is from $\mathbf{M_{\mathcal{X}\mapsto\mathcal{Y}}}(x)$, SCALE essentially conduct few-shot learning in a multi-task way by introducing an additional refinement task. Refinement has long been proved effective in MT~\citep{xia2017deliberation,cheng2022neural}. And this also holds true for LLM-based translation. In this refinement process, we pass sampled sentences and their confidence score (probability score) from STM to an LLM. The LLM then digests the information carried by the sampled set and infers the generation space of the STM, which guides the LLM to generate the output that is more consistent with the local data distribution~\citep{xu2023small}. And since the final translation is delivered by an LLM, SCALE could also mitigate the parallel data bias from STMs and exhibit robustness by not merely copying and pasting the draft translation from STMs as shown in \S\ref{section:ablation}.

\textbf{Pivoting}
Considering the predominantly English-centric nature of most LLMs~\citep{DBLP:journals/corr/abs-2005-14165,touvron2023llama}, SCALE could employ an intermediate variable $\mathbb{Z}$ from $\mathbf{M_{\mathcal{X}\mapsto\text{English}}}(x)$ where the target language $\mathcal{Y}$ is not necessarily English. And here $\mathbb{Z}$ serves as a pivot point for LLMs to enhance their understanding of the source sentence and yield improved translations. This can also be regarded as a form of knowledge transfer from high-resource languages to low-resource languages~\citep{chen2017teacher,kim2019pivot,jiao2023chatgpt}.

\textbf{Updating}  
A significant limitation of the existing LLM-based translation systems is the inherent complexity of LLM continual learning. This complexity arises from several factors, including the delicate balance between speciality and generality \citep{lin2023speciality}, the catastrophic forgetting problem \citep{yong2023bloom1}, and the substantial computational demands \citep{dettmers2023qlora}. In contrast, the SCALE framework offers a more efficient and streamlined approach to continuous updating. By exclusively and effectively updating the lightweight $\mathbf{M_{\mathcal{X}\mapsto\cdot}}$ component, the framework ensures that the LLM remains untouched, thus preserving its general language capabilities. This selective updating process not only mitigates the issue of catastrophic forgetting but also reduces the computational burden of fine-tuning associated with LLM-based translation systems.

\section{Experimental Setup}
\subsection{Dataset}
Our evaluation datasets encompass a diverse set of languages, spanning both low- and high-resource settings and deriving from various language families. To facilitate reproducibility and data sharing, all our evaluation datasets come from the \texttt{devtest} split of Flores-200~\citep{nllb2022}, a publicly available many-to-many evaluation data set covering 200 languages from all over the world.

\subsection{Translation Systems}
\label{section:translation_systems}
We compare our approach with cutting-edge academic systems including both specialized models and LLMs, as well as one commercial system, Microsoft Translator\footnote{\url{https://azure.microsoft.com/en-us/products/cognitive-services/translator}}.  

We have two strong specialized models:
\begin{itemize}
    \item \textbf{M2M100}~\citep{DBLP:journals/jmlr/FanBSMEGBCWCGBL21} is the first multilingual encoder-decoder translation model that can translate between any pair of 100 languages without relying on English data.
    \item \textbf{NLLB}~\citep{nllb2022} is a supervised translation model suite covering from 169M to 54.5B~(MOE) parameters with encoder-decoder architecture and capable of delivering high-quality translations directly between 200 languages.
\end{itemize}
For few-shot LLMs, we consider:
\begin{itemize}
    \item \textbf{XGLM}~\citep{DBLP:conf/emnlp/LinMAWCSOGBDPSK22} is a multilingual generative language models trained on a corpus covering a diverse set of languages and the largest XGLM-7.5B model outperforms comparable sized GPT-3 model in multilingual setting.
    \item \textbf{GPT-3.5}\footnote{\url{https://platform.openai.com/docs/models/gpt-3-5}} is a GPT model specially optimized for conversational purpose and shows remarkable performance in machine translation tasks~\citep{jiao2023chatgpt}.
    \item \textbf{GPT-4} \citep{OpenAI2023GPT4TR} is the latest and the most powerful version of GPT-series.
\end{itemize}
We use both GPT-3.5 and GPT-4 from Microsoft Azure OpenAI Service\footnote{\url{https://azure.microsoft.com/en-us/products/ai-services/openai-service}}. Without further notice, the number of few-shot samples in LLM and SCALE are set to 10 and the sample selection strategy follows ~\citet{agrawal2022incontext}. The prompt we use could be found in the Appendix~\ref{appendix:prompt}.

\subsection{Evaluation Metrics}
Because neural metrics have shown higher correlation with human preference~\citep{DBLP:conf/wmt/FreitagRMLSAKFLM22,DBLP:conf/emnlp/ReiSFL20} and are widely adopted by recent literatures~\citep{DBLP:journals/corr/abs-2302-09210,DBLP:journals/corr/abs-2302-01398}, we mainly evaluate our system with (1) \textbf{COMET-22}\footnote{\url{https://huggingface.co/Unbabel/wmt22-comet-da}}, a reference-based neural metric~\citep{DBLP:conf/wmt/ReiSAZFGLCM22} combining direct assessments, sentence-level score, and word-level tags from multidimensional quality metrics error annotations, (2) \textbf{COMETKiwi}~\footnote{ \url{https://huggingface.co/Unbabel/wmt22-cometkiwi-da}}, a refrence-free quality estimation model from ~\citet{rei-etal-2022-cometkiwi}, and (3) \textbf{\textsc{Bleurt}}~\citep{DBLP:conf/acl/SellamDP20}, a learnable evaluation metric with a regression model trained on ratings data. 
For completeness, we also include the results of lexical metrics such as spBLEU~\citep{nllb2022} and chrF++~\citep{DBLP:conf/wmt/Popovic17}.

\section{Experimental Results}
In this section, we conduct various experiments to show the effectiveness of our framework. In \S\ref{sectioin:refinement}, we verify the effectiveness of the refinement ability within SCALE by comparing with STMs and few-shot LLMs. In \S\ref{sectioin:pivoting}, we focus on non-English pairs to test the pivoting ability of SCALE. In \S\ref{sectioin:update}, we show the continual learning results of SCALE with a fixed LLM and an evolving STM. 

\subsection{SCALE Refinement}
\label{sectioin:refinement}

To evaluate the refinement capabilities of SCALE, this section primarily concentrates on low-resource languages, which currently pose significant challenges for few-shot LLMs. Our approach showcases its versatility by incorporating languages from diverse families and scripts, including Assamese (asm\_Beng), Armenian (hye\_Armn), Amharic (amh\_Ethi), Xhosa (xho\_Latn), Uyghur (uig\_Arab), Khmer (khm\_Khmr), Nepali (npi\_Deva), and Sindhi (snd\_Arab). For additional data details, please refer to the Appendix~\ref{appendix:data_statistics}.
\input{tables/refinement_lowres}
We adopt three kinds of baseline systems as described in \S\ref{section:translation_systems}. For supervised NLLB model suite, we choose the NLLB-3.3B version, and for SCALE-refine, the LLM is GPT-4 the STM is also NLLB-3.3B for fair comparison.

The results are displayed in Table~\ref{table:refinement_lowres}. As observed, few-shot LLMs, including GPT-4, significantly trail behind specialized models in all translation directions. Even with Xhosa belonging to the same language family as English, the GPT-4 model fails to deliver comparable results to NLLB model. In contrast, our framework, by combining LLMs and STMs, demonstrates superior performance over few-shot GPT-4 by an averaged 2.96 COMET scores and 5 BLEURT scores, and surpasses the strong NLLB model in 8/8 directions. Interestingly, when the performance gap is substantial (e.g., SCALE-refine over GPT-4), the lexical metric spBLEU aligns with COMET and BLEURT. However, when comparing SCALE-refine with NLLB, although COMET-22, COMETkiwi, and BLEURT exhibit consistent patterns, spBLEU displays degradation with the GPT-based system in 4 out of 8 directions. Similar findings are also reported in \citet{vilar2023prompting,DBLP:journals/corr/abs-2302-09210}.

\begin{figure*}[t!]
  \centering
  \includegraphics[width=0.95\textwidth,height=0.27\textwidth]{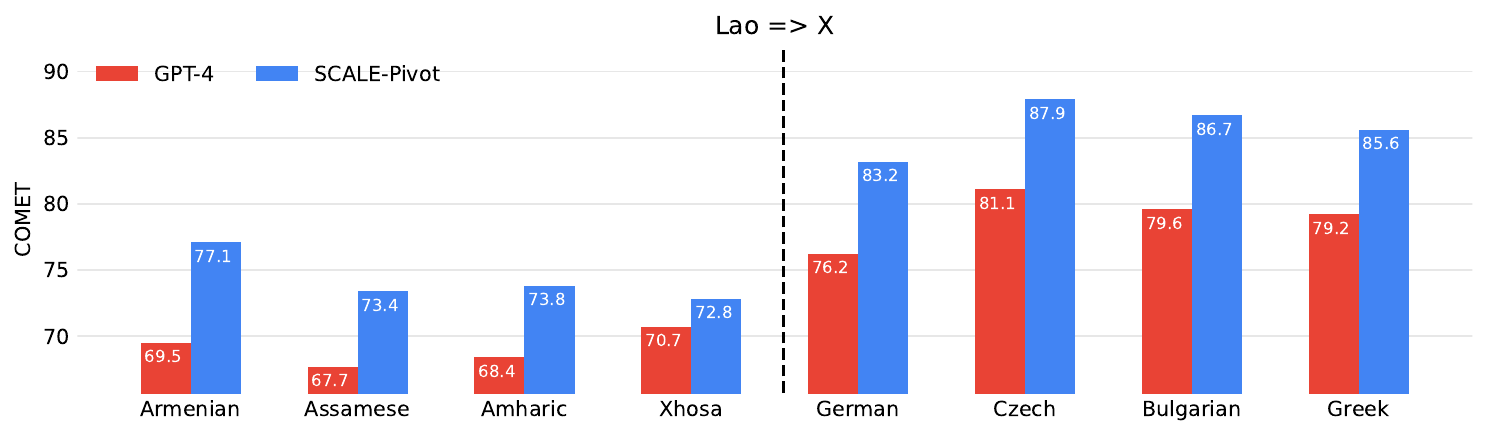}
  \caption{Translation results from Lao to both low- and high-resource languages, where GPT-4 uses few-shot prompting and SCALE-pivot uses English as the pivot language.}
  \label{figure:pivot}
\end{figure*}
\subsection{SCALE Pivoting}
\label{sectioin:pivoting}

In this section, we demonstrate the performance of SCALE-pivot, in which the variable $\mathbb{Z}$ is not directly pertinent to the current translation directions but functions as a pivot. Specifically, we examine the performance of few-shot GPT-4 and SCALE-pivot on Lao$ \rightarrow\mathbb{Y}$ translations, where $\mathbb{Y}$ represents a language set encompassing both low-resource and high-resource languages. For the low-resource languages, we include Assamese (asm\_Beng), Armenian (hye\_Armn), Amharic (amh\_Ethi), Xhosa (xho\_Latn), and we have German (deu\_Latn), Czech (ces\_Latn), Bulgarian (bul\_Cyrl) and Greek (ell\_Grek) for the high-resource setting. 

The results are presented in Figure~\ref{figure:pivot}. Firstly, with GPT-4 results alone, we could observe that the language bias of LLM heavily affects translation performance. The few-shot GPT-4 model typically excels in the high-resource setting but struggles in low-resource one. Furthermore, it is evident that SCALE-pivot can enhance the performance of GPT-4 in both low- and high-resource settings, while the performance gain is more significant in high-resource setting (an averaged 6.8 COMET-22 score improvement for high-resource versus 5.2 for low-resource).

\subsection{SCALE Updating}
\label{sectioin:update}
\begin{figure*}[thb!]
  \centering
  \includegraphics[width=0.9\textwidth,height=0.64\textwidth]{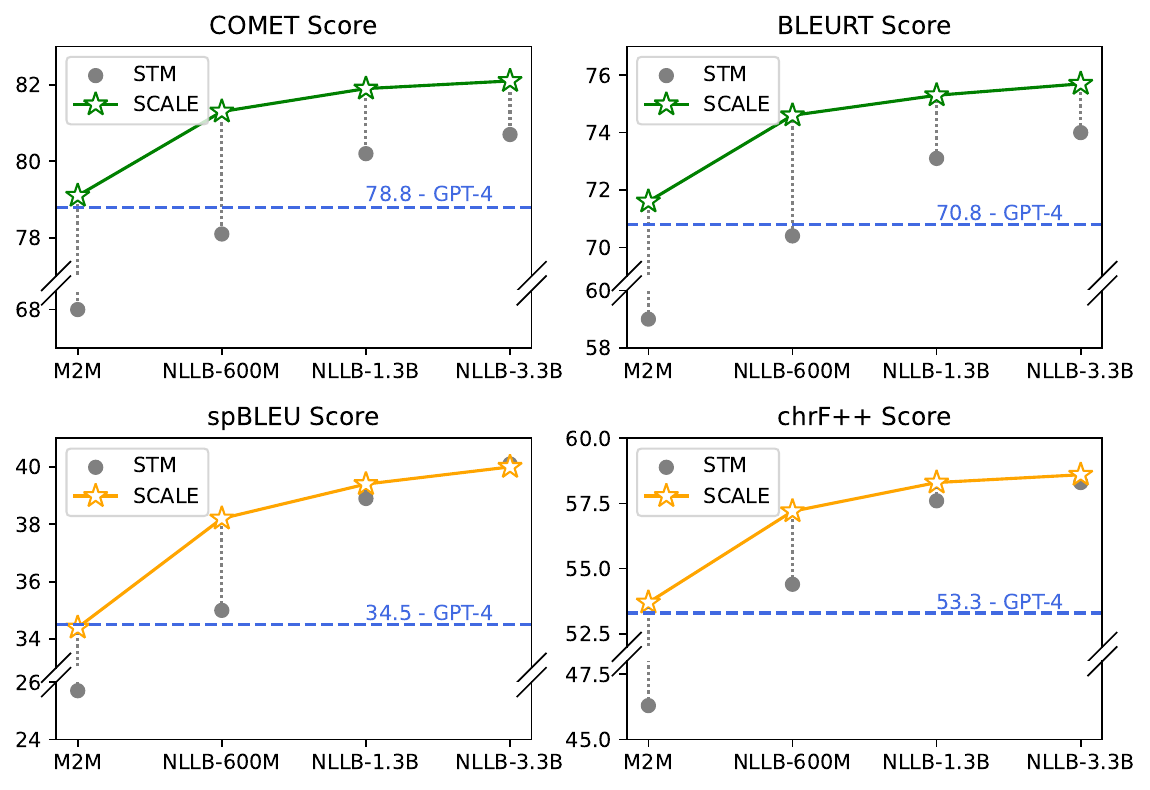}
  \caption{Translation results from Xhosa to English with evolving STMs in the SCALE framework.}
  \label{figure:update}
\end{figure*}

In this section, we explore the potential enhancement of our framework by keeping the LLM fixed and solely updating the STM. 
Specifically, we use M2M100-12B and NLLB model suite ranging from 600M to 3.3B as our evolving STM.
We conduct experiments on the Xhosa $\rightarrow$ English direction and adopt the prompt format of SCALE-refine.
The experimental results are displayed in Figure~\ref{figure:update}, leading to the following observations: 

(1) The overall framework can be consistently improved with a fixed LLM and a continuously evolving STM; 
(2) SCALE, when equipped with a small model containing only 600M parameters, can outperform GPT-4 with an absolute 2.5 COMET-22 score and a 3.8 BLEURT score; 
(3) Equipped with an STM (M2M100) of relatively lower performance than original few-shot GPT-4 , SCALE demonstrates strong robustness by not merely copying and pasting the less satisfactory reference answer provided by M2M100, which we detailedly investigated in $\S$\ref{section:ablation}.

Interestingly, we also observe that the growth patterns exhibited by lexical metrics and neural semantic metrics differ. For M2M100 and NLLB-600M as STM, both metrics experience substantial improvement, while for NLLB-1.3B and 3.3B as STM, SCALE maintains the same lexical accuracy while continually enhancing translation performance as measured by neural semantic metrics.

\begin{figure*}[t!]
  \centering
  \includegraphics[width=0.9\textwidth,height=0.27\textwidth]{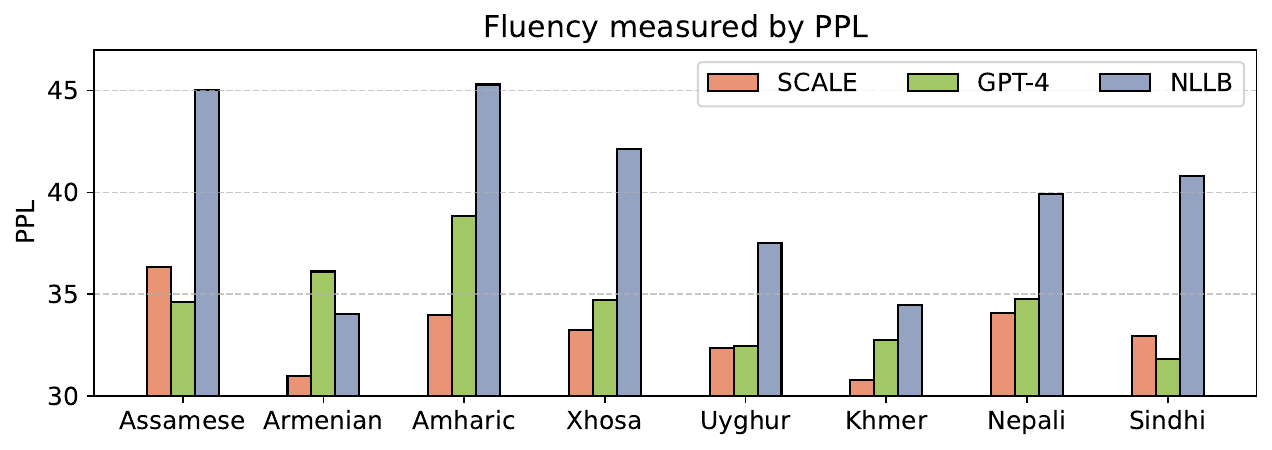}
  \caption{Perplexity score from $\mathbb{X}$$\rightarrow$English translation measured by GPT2-XL.}
    \label{figure:ppl}
\end{figure*}
\section{Further Analysis}
\subsection{Translation Characteristics}
\label{section:translation_characteristics}

To gain a deeper understanding of the translation characteristics of different systems (few-shot LLMs, STMs, and SCALE) beyond overall translation quality, we employ the following measurements, as suggested by \citet{DBLP:journals/corr/abs-2302-09210}:

\begin{enumerate}
\item \textbf{Translation Fluency:} Since LLMs are optimized by predicting the next token, their translations tend to display a language modeling bias that favors fluency over adequacy. To investigate this, we utilize an independently trained open-source language model (GPT2-XL~\citep{Radford2019LanguageMA}) to measure the perplexity score of the translation output.

\item \textbf{Translation Non-Monotonicity:} This metric evaluates the extent to which a translation adheres to the source sentence's structure, calculating the deviation from the diagonal in the word-to-word alignment. Translations that are more paraphrastic or less literal tend to deviate from closely tracking the source word order across language pairs~\citep{DBLP:journals/corr/abs-2302-09210}. We apply the non-monotonicity metric proposed by \citet{schioppa-etal-2021-controlling}.

\item \textbf{Unaligned Source Words:} Another measure of literalness is the count of unaligned source words~\citep{DBLP:journals/corr/abs-2302-09210,raunak2023gpts}. When accounting for quality, less literal translations are likely to include more words that do not align with those in the source sentence. 
\end{enumerate}

We present the \textbf{Translation Fluency} results of $\mathbb{X}\rightarrow$ English translation in Figure~\ref{figure:ppl}, where $\mathbb{X}$ remains the same as used in Section \ref{sectioin:refinement}. It is evident that regardless of the translation quality delivered by the LLM, whether superior (SCALE) or inferior (GPT-4) compared to the STM (NLLB), the LLM translation generally demonstrates higher fluency than the STM. Additionally, in 6 out of the 8 languages examined, SCALE produces lower perplexity scores than the original GPT-4 output. This suggests that the STM-generated variable $\mathbb{Z}$ can effectively aid the GPT-4 model in further decreasing its generation uncertainty.

\begin{figure*}[htb!]
  \centering
  \includegraphics[width=0.9\textwidth,height=0.32\textwidth]{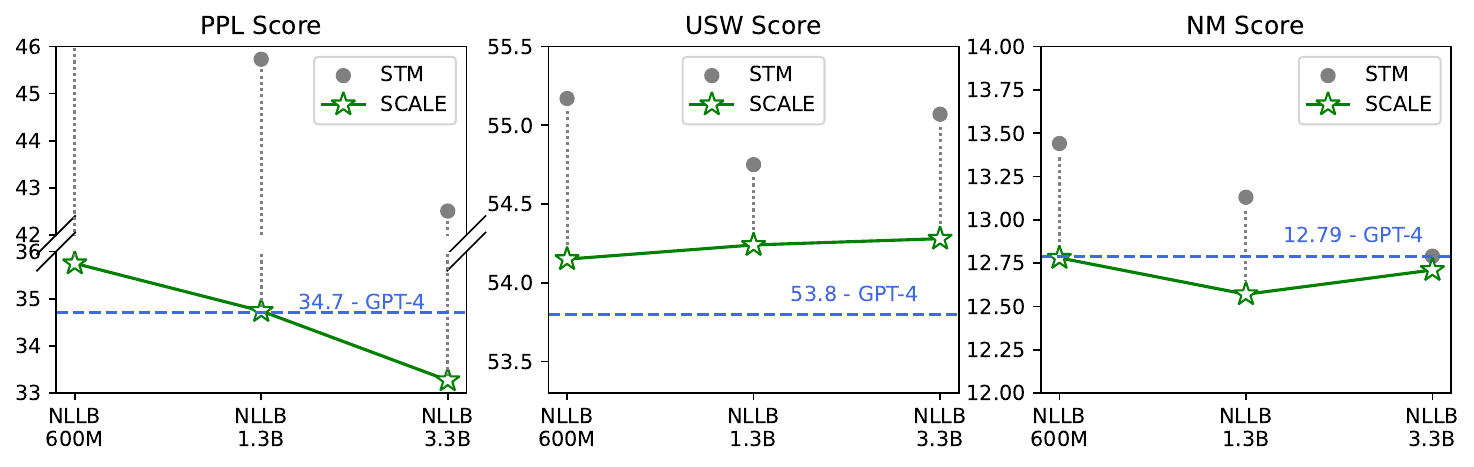}

  \caption{Perplexity, Unaligned Source Words percentage and Non-Monotonicity score from Xhosa$\rightarrow$English translation.}
    \label{figure:analysis}
\end{figure*}

For \textbf{Non-Monotonicity} and \textbf{Unaligned Source Words}~\footnote{We use this implementation: \url{ https://github.com/vyraun/literalness} with \textit{xlmr} branch in \url{https://github.com/neulab/awesome-align/tree/xlmr}}, we choose Xhosa$\rightarrow$English translation with different STMs, and the results are shown in Figure~\ref{figure:analysis}. We also include PPL score for completeness. We find that both the USW and NM scores for STM are higher than those of GPT-4. This indicates that even though STM provides higher translation quality, it results in less literal translations. However, for SCALE, it effectively reduces GPT-4's NM score while maintaining a moderate USW score. This suggests that during the SCALE refinement process, the model primarily adheres to the original LLM output structure while taking cues from STM's word selection. We show several concrete cases in Appendix \ref{appendix:case}.

\subsection{Multipath Sampling}
\label{section:multipath}
\input{tables/multipath_sampling}
In this section, We list the results of multiple path sampling strategy in Table~\ref{table:multipath}. We test with Xhosa$\rightarrow$English with one-shot SCALE-refine. The results show that without increasing the shot number in the few-shot learning, using STM to generate more generation paths could consistently improve the overall performance, which could be useful in the extremely low-resource setting where demonstration samples are hard to acquire.

\subsection{Ablation}
\label{section:ablation}
In this section, we conduct an ablation study for each key design in our framework. We examine the following variants: (1) without confidence: This model follows the same setting as the SCALE-refine in \S\ref{sectioin:refinement}, except that we do not pass the confidence score of each token as input. (2) zero-shot: This variant removes all in-context-learning examples, keeping only the translation instruction and the reference answer from STM. (3) one-shot: This model utilizes only one-shot, in contrast to the ten-shot results presented in \S\ref{sectioin:refinement}. (4) zero-shot-M2M: This model also implements zero-shot, but the STM used is M2M100, a less performant model than the original few-shot GPT-4. This is employed to assess the robustness of our framework.

The outcomes of our ablation study are showcased in Table~\ref{table:ablation}. It is evident that each component in our framework perform effectively, with the in-context-learning setting providing the most performance gain. This indicates that simply offering a reference answer to the LLM without in-context samples does not adequately guide the model in utilizing those references effectively. Furthermore, the number of ICL examples is also an essential factor in the process.

Regarding the SCALE zero-shot-M2M variant, its performance is significantly inferior to that of the few-shot LLM due to the poor quality of the M2M100 output. From this observation, we can conclude that the robustness of SCALE, as illustrated in Figure~\ref{figure:update}, primarily stems from the power of in-context learning. This learning approach informs the LLM about which elements to trust and which to disregard, ultimately improving the overall translation performance and robustness.
\input{tables/ablation}
\subsection{Generation Latency}
\label{section:latency}
\input{tables/latency}
In this section, we conduct a detailed evaluation of the overhead introduced by SCALE in comparison to conventional few-shot LLM. The additional latency arises from two factors: first, the time required to generate the variable $\mathbb{Z}$ for the current source sentence $x$ using STM, and second, the increased latency caused by the LLM due to the extended context. Since the response time from the GPT API may not accurately represent the actual latency of the LLM, we utilize one of the largest open-source LLMs (BLOOM-176B) for this analysis. As shown in Table~\ref{table:latency}, we observe that the incurred latency can be primarily attributed to the extended context window due to the quadratic time complexity of the transformer architecture. Exploring methods to accelerate this process based on STM-generated output using speculative decoding techniques remains a topic for future work \citep{xia2022speculative, chen2023accelerating, yang2023inference}.

\section{Related Work}
The use of LLM for translation tasks has garnered significant interest in recent times. ~\citet{DBLP:journals/corr/abs-2005-14165} initially demonstrated the efficacy of prompting an LLM with a few examples to achieve noteworthy results, particularly in high-resource languages~\citep{vilar2023prompting, DBLP:conf/emnlp/LinMAWCSOGBDPSK22}. Following the release of ChatGPT, several studies have examined its overall translation performance\citep{jiao2023chatgpt, DBLP:journals/corr/abs-2302-09210}, along with works focusing on the issue of hallucination~\citep{guerreiro2023hallucinations} , literalness~\citep{raunak2023gpts}, multilinguality~\citep{zhu2023multilingual} and incidental bilingualism problem~\citep{briakou2023searching}. A comprehensive analysis conducted by \citet{DBLP:journals/corr/abs-2302-01398} revealed the unreasonable effectiveness of few-shot LLMs. Furthermore, a diverse range of research has attempted to enhance LLM-based translation systems through cultural awareness \citep{yao2023empowering}, refinement \citep{chen2023iterative,cheng2023lift}, retrieval-augmentation \citep{cheng2023lift}, post-editing \citep{raunak2023leveraging}, and comparison \citep{zeng2023tim}. 

Our work also shares similarities with a series of studies that aim to build collaboration between LLMs and other systems. \citet{luo2023augmented} propose equipping LLMs with a knowledge-guiding module to access relevant information without altering the LLMs' parameters. \citet{DBLP:journals/corr/abs-2302-09210} propose to use Microsoft Translator system as the primary translation system, and then use GPT as a fallback system when the quality of MS-Translator is unsatisfactory measured by reference-free metrics. 
\citet{xu2023small} introduce SuperICL and achieve significant improvements in various language understanding tasks. 
\citet{ge2023context} employ a trainable LoRA-based encoder as an additional model for LLM context compression. 

\section{Conclusion}
In this paper, we present a novel collaborative framework SCALE, which effectively combines the strengths of Large Language Models (LLMs) and compact Specialized Translation Models (STMs) through an in-context learning approach. By providing triplet in-context demonstrations, our framework successfully unlocks the refinement and pivoting capabilities of LLMs. SCALE demonstrates its superiority in many scenarios including low-resource setting, multilingual translation and model continual learning setting. Our results offer crucial understanding and a robust basis for subsequent research investigating the possible synergistic effects between LLMs and more specialized models tailored for specific tasks.

\subsubsection*{Acknowledgments}
We would like to acknowledge Jiduan Liu and Lemao Liu for the helpful discussions and valuable suggestions.

\bibliography{iclr2024_conference}
\bibliographystyle{iclr2024_conference}

\clearpage
\appendix
\section{Appendix}

\subsection{Prompt Example}
\label{appendix:prompt}
In Table~\ref{table:prompt_llm}, we list the prompt we use for few-shot LLM and in Table~\ref{table:prompt_SCALE}, for our SCALE framework. We use Chat Markup Language version from Azure to format our prompt\footnote{\url{https://learn.microsoft.com/en-us/azure/ai-services/openai/how-to/chatgpt?pivots=programming-language-chat-ml}}.

\input{tables/llm_prompt}

\input{tables/scale_prompt}

\subsection{Data Statistics}
\label{appendix:data_statistics}
We list the detailed data information for SCALE-refine and SCALE-Pivot experiments in Table~\ref{table:data_statistics}. The number of dev set is 997 and 1012 for devtest set in flores-200~\citep{nllb2022}.
\input{tables/statistics}

\subsection{Translation Cases}
\label{appendix:case}
In this section, we list several translation cases from different languages.

\begin{figure*}[ht]
\centering
\includegraphics{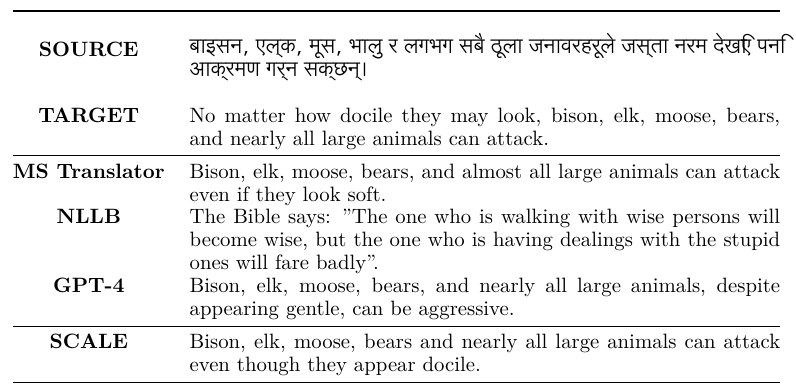}
  \caption{Translation case from Nepali to English.}
    \label{figure:case_1}
\end{figure*}

\begin{figure*}[ht]
\centering
\includegraphics{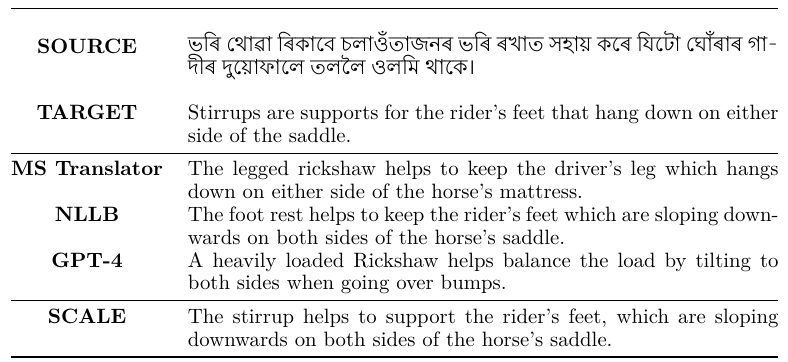}
  \caption{Translation case from Assamese to English.}
    \label{figure:case_2}
\end{figure*}

\begin{figure*}[ht]
\centering
\includegraphics{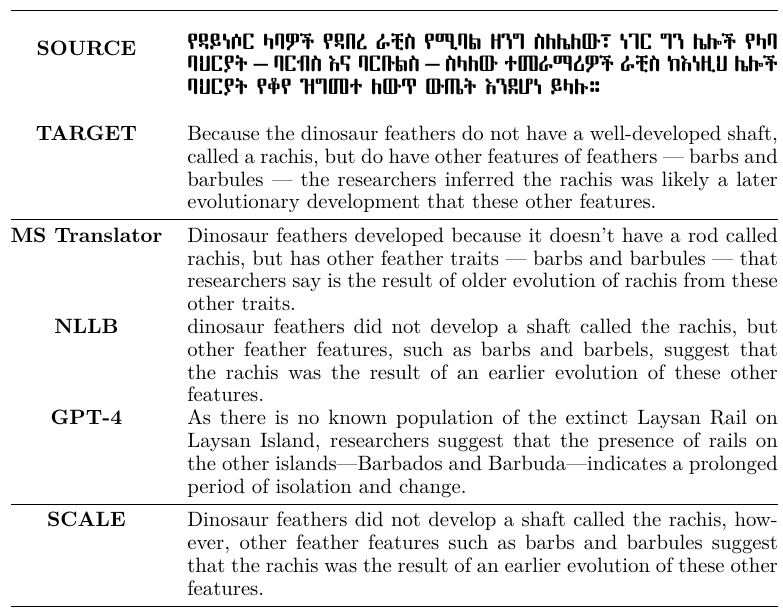}
  \caption{Translation case from Amharic to English.}
    \label{figure:case_3}
\end{figure*}

\begin{figure*}[ht]
\centering
\includegraphics{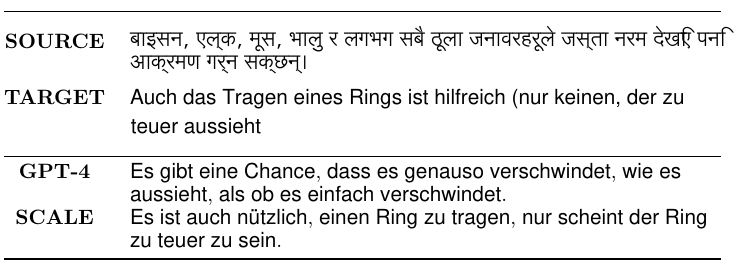}
  \caption{Translation case from Lao to German.}
    \label{figure:case_4}
\end{figure*}
\end{document}

%% file: math_commands.tex
\usepackage{amsmath,amsfonts,bm}

\def\eqref#1{equation~\ref{#1}}

\def\1{\bm{1}}

\DeclareMathAlphabet{\mathsfit}{\encodingdefault}{\sfdefault}{m}{sl}
\SetMathAlphabet{\mathsfit}{bold}{\encodingdefault}{\sfdefault}{bx}{n}



%% file: tables/refinement_lowres.tex
\begin{table*}[htbp!]
\centering
\resizebox{\linewidth}{!}{
\begin{tabular}{lcccc|cccc}
\toprule
& \bf COMET-22 & \bf COMETKiwi & \textbf{BLEURT} & \bf spBLEU & \bf COMET-22 & \bf  COMETKiwi & \bf BLEURT  & \bf spBLEU\\ \hline
& \multicolumn{4}{c|}{asm\_Beng } & \multicolumn{4}{c}{hye\_Armn }  \\
NLLB   & \underline{85.6} & \underline{82.8}  & \underline{72.1}  & 33.9 & \underline{88.3}  & \underline{87.5} & \underline{77.0} & 43.0 \\
M2M100 & n/a & n/a & n/a & n/a &75.9&	76.5&	58.9&	23.7 \\
Microsoft & 83.5 & 81.7  & 68.8  & 29.6 & 85.2  &  85.0 & 71.5 & 34.6  \\
XGLM & 62.7	& 57.8	& 38.8	& 3.7 &	43.9 & 50.2 & 20.5 & 0.2  \\
GPT-3.5 & 78.6 & 76.7  & 61.0  & 18.1 & 77.0  & 77.2 & 60.5 & 19.4  \\
GPT-4 & 83.9 & 80.9 & 69.1 & 27.9 & 86.2 & 86.0 & 73.1 & 35.6  \\
SCALE-refine  & \bf 86.6 &\bf 	83.2 & \bf 73.8 & 34.1 &\bf  88.8 &\bf  88.0 &\bf  77.8 & 42.3 \\
\hline
& \multicolumn{4}{c|}{amh\_Ethi } & \multicolumn{4}{c}{xho\_Latn }  \\
NLLB  & 86.9 & 84.5 & 73.6 & 36.4 & \underline{80.7} & \underline{65.8} & \underline{74.0} & 40.1  \\
M2M100 &72.3&	72.0	&54.8&	18.5&	68.0&	62.1&	59.0	&25.7 \\
Microsoft & \underline{87.5} & \underline{84.6}  & \underline{74.7}  & 41.9 & n/a  & n/a & n/a & n/a \\
XGLM &50.2	& 43.9	&17.8	&0.1&	39.6&	41.7&	37.1&	1.6   \\
GPT-3.5 & 58.8 & 54.2  & 31.7  & 3.4 & 69.1  & 65.5 & 58.3 & 21.9  \\
GPT-4  & 83.2 & 81.9 & 67.3 & 27.1 & 78.8 & 67.1 & 70.8 & 34.5 \\
SCALE-refine  & \bf 88.0 &\bf  85.3 &\bf  75.7 & 37.6 &\bf  82.1 &\bf  67.3 &\bf  75.7 & 40.0  \\
\hline
& \multicolumn{4}{c|}{uig\_Arab } & \multicolumn{4}{c}{khm\_Khmr }  \\
NLLB  & \underline{85.4} & \underline{84.4} & \underline{70.4} & 27.5 & \underline{86.1} & \underline{85.4} & \underline{72.2} & 35.4  \\
M2M100 &n/a&n/a&n/a&n/a& 	69.6&	71.6&	54.0&	17.6 \\
Microsoft & 82.7 & 81.7  & 66.2  & 21.6 & 80.2  & 80.5 & 63.3 & 25.6  \\
XGLM & 37.1	& 52.8	& 16.9& 	0.2	& 48.6& 	53.7&	21.6&	0.7   \\
GPT-3.5 & 73.7 & 74.2  & 53.0  & 11.6 & 73.3  & 73.0 & 53.2 & 13.9  \\
GPT-4  & 83.7 & 82.8 & 67.4 & 23.1 & 84.6 & 84.0 & 69.9 & 29.1  \\
SCALE-refine  &\bf  86.4 &\bf  85.0 &\bf  72.2 & 27.9 &\bf  87.1 & \bf 85.9 &\bf  73.9 & 34.7   \\
\hline
& \multicolumn{4}{c|}{npi\_Deva } & \multicolumn{4}{c}{snd\_Arab }  \\
NLLB  & \underline{90.4} & \underline{88.3} & \underline{77.1} & 45.0 & \underline{86.9} & \underline{79.5} & \underline{75.5} & 44.4  \\
M2M100 & 75.2&	73.6&	55.1&	21.2&	49.8&	47.2&	39.2&	6.4 \\
Microsoft & 89.8 & 88.2  & 75.3  & 42.8 & 83.6  & 77.4 & 70.4 & 38.5  \\
XGLM &72.9	&67.0	&48.8	&8.3	&53.8&	45.1	&29.8&	1.8   \\
GPT-3.5 & 87.2 & 85.4  & 69.9  & 29.3 & 75.6  & 68.1 & 58.8 & 17.3  \\
GPT-4  & 90.2 & 88.1 & 76.3 & 40.8 & 83.2 & 75.3 & 69.9 & 32.3  \\
SCALE-refine  & \bf 91.1 & \bf 88.8 &\bf  78.1 & 44.0 & \bf 87.5 &\bf  79.5 &\bf  76.6 & 42.9   \\
\hline
\end{tabular}
}

\caption{Translation results of eight low-resource languages to English. The best results are in \textbf{bold} and the second best are with \underline{underscore}. SCALE-refine is compared with specialized model (NLLB, M2M), commercial system (MS Translator) and few-shot LLM (XGLM, GPT-3.5, GPT-4).}
\label{table:refinement_lowres}
\end{table*}

%% file: tables/multipath_sampling.tex
\begin{table*}[t!]
\centering
\begin{tabular}{cccc}
\toprule
\# Path & \bf COMET-22 & \textbf{BLEURT} & \bf spBLEU \\ \hline
1 & 80.4	&73.2	& 35.6 \\
2 & 81.2	&74.3	& 37.1 \\
3 & 81.4	&74.7	& 38.0 \\
4 & 81.5	&74.8	& 38.3 \\
5 & 81.4	&74.9	& 38.4 \\
\hline
\end{tabular}

\caption{Translation results from Xhosa to English with multi-path sampling. All the experiments are conducted by one-shot SCALE-refine and only differ in the number of sampled paths from STM.}
\label{table:multipath}
\end{table*}

%% file: tables/ablation.tex
\begin{table*}[htbp!]
\centering
\begin{tabular}{lccc}
\toprule
& \bf COMET-22 & \bf COMETKiwi & \textbf{BLEURT} \\ \hline
M2M100   & 68.0	& 62.1 &	59.0 \\
NLLB   & 80.7	& 65.8 &	74.0 \\
GPT-4 &  78.8 & 67.1 &  70.8  \\
\hdashline
SCALE & 82.1&	67.3&	75.7 \\
 \quad w/o confidence & 81.6	&67.6	&74.9\\
 \quad zero-shot &81.4 & 66.4 & 74.8\\
 \quad one-shot & 81.7 & 66.7 & 75.3 \\
 \quad zero-shot-M2M & 76.4 &66.8 & 68.2 \\
\hline
\end{tabular}
\caption{Ablation study for SCALE with Xhosa$\rightarrow$English translation.}
\label{table:ablation}
\end{table*}

%% file: tables/latency.tex
\begin{table*}[thbp!]
\centering
\begin{tabular}{lcc|cccc}
\toprule
& \multicolumn{2}{c|}{\textbf{few-shot LLM}}  & \multicolumn{4}{c}{\textbf{SCALE}} \\ 
  & avg. \#length	& total & avg. \#length	& STM & LLM & total \\ \hline
0-shot & 101.37 & 7.19 & 161.13 & 1.87 & 7.43 & 9.3\\
1-shot & 198.00 & 7.46 & 516.92 & 1.87 & 8.33 & 10.2\\
10-shot & 951.91 & 9.52 & 2489.72 & 1.87 & 14.17 & 16.04\\
\hline
\end{tabular}
\caption{Generation latency results of LLM (BLOOM-175B) and SCALE (BLOOM-175B + NLLB-3.3B) measured in seconds (s).}
\label{table:latency}
\end{table*}

%% file: tables/llm_prompt.tex
\begin{table}[htbp!]
\centering
\begin{tabular}{ll}
\toprule
\multirow{4}{*}{Instruction} & \imstart system\\
            &Assistant is an intelligent chatbot designed \\
            & to help users translate from \source{\$\{source\_language\}} to \target{\$\{target\_language\}} \\
            & \imend \\
\midrule
\multirow{6}{*}{Examples} &\imstart user\\
           &Source: \source{\$\{source\_1\}} \\
           & Target: \target{\$\{target\_1\}} \\
           & ... \\
           & Source: \source{\$\{source\_n\}} \\
           & Target: \target{\$\{target\_n\}} \\
\midrule
\multirow{4}{*}{Input}    & Source: \source{\$\{source\}} \\
           &\imend \\
           &\imstart assistant \\
           &Target:  \\
\bottomrule
\end{tabular}
    \caption{Prompt of Chat Markup Language format for few-shot LLM.}
    \label{table:prompt_llm}
\end{table}

%% file: tables/scale_prompt.tex
\begin{table}[htbp!]
\centering
\begin{tabular}{ll}
\toprule
\multirow{10}{*}{Instruction} & \imstart system\\
            &Assistant is an intelligent chatbot designed \\
            & to help users translate from \source{\$\{source\_language\}} to \target{\$\{target\_language\}} \\
            & \\
            & Context: \\
            & $\cdot$ Assistant would would be given a potentially useful \reference{reference } answer \\
            &  from a fine-tuned model \\
            & $\cdot$ The number in brackets denotes the confidence score of a fine-tuned model \\
            &  to generate the token.\\
            & \imend \\
\midrule
\multirow{8}{*}{Examples} &\imstart user\\
           &Source: \source{\$\{source\_1\}} \\
           &Potentially useful reference answer 1: \reference{\$\{reference\_1\}} \\
           &Potentially useful reference answer 2: \reference{\$\{reference\_2\}}\\
           & Target: \target{\$\{target\_1\}} \\
           & ... \\
           & Source: \source{\$\{source\_n\}} \\
           &Potentially useful reference answer 1: \reference{\$\{reference\_1\}} \\
           &Potentially useful reference answer 2: \reference{\$\{reference\_2\}}\\
           & Target: \target{\$\{target\_n\}} \\
\midrule
\multirow{5}{*}{Input}    & Source: \source{\$\{source\}} \\
&Potentially useful reference answer 1: \reference{\$\{reference\_1\}} \\
           &Potentially useful reference answer 2: \reference{\$\{reference\_2\}}\\
           &\imend \\
           &\imstart assistant \\
           &Target:  \\
\bottomrule
\end{tabular}
    \caption{Prompt of Chat Markup Language format for SCALE.}
    \label{table:prompt_SCALE}
\end{table}

%% file: tables/statistics.tex
\begin{table}[htbp!]
\resizebox{\linewidth}{!}
{
\begin{tabular}{@{}ccccccc@{}}
\toprule
\textbf{code}                 & \textbf{language}           & \textbf{\# dev length}    & \textbf{\# devtest length} & \textbf{script}              & \textbf{family}                   & \textbf{resource}        \\ \midrule
asm\_Beng                     & Assamese                    & 40.55                     & 41.67                      & Bengali                      & Indo-European                     & low                      \\
hye\_Armn                     & Armenian                    & 43.91                     & 45.31                      & Armenian                     & Indo-European                     & low                      \\
amh\_Ethi                     & Amharic                     & 38.87                     & 39.64                      & Ge'ez                        & Afro-Asiatic                      & low                      \\
xho\_Latn                     & Xhosa                       & 35.31                     & 36.37                      & Latin                        & Atlantic-Congo                    & low                      \\
uig\_Arab                     & Uyghur                      & 40.77                     & 42.41                      & Arabic                       & Turkic                            & low                      \\
khm\_Khmr                     & Khmer                       & 52.77                     & 53.79                      & Khmer                        & Austroasiatic                     & low                      \\
npi\_Deva                     & Nepali                      & 34.36                     & 35.48                      & Devanagari                   & Indo-European                     & low                      \\
eng\_Latn                     & English                     & 28.99                     & 30.28                      & Latin                        & Indo-European                     & high                     \\
deu\_Latn                     & German                      & 37.57                     & 39.16                      & Latin                        & Indo-European                     & high                     \\
ces\_Latn                     & Czech                       & 36.63                     & 38.10                      & Latin                        & Indo-European                     & high                     \\
bul\_Cyrl                     & Bulgarian                   & 37.99                     & 39.45                      & Cyrillic                     & Indo-European                     & high                     \\
rus\_Cyrl & Russian& 39.42 & 40.21  & Cyrillic & Indo-European & high \\
\bottomrule
\end{tabular}
}
\label{table:data_statistics}
\caption{Data statistics for all the tested languages in the paper.}

\end{table}

%% file: main.bbl
\begin{thebibliography}{52}
\providecommand{\natexlab}[1]{#1}
\providecommand{\url}[1]{\texttt{#1}}
\expandafter\ifx\csname urlstyle\endcsname\relax
  \providecommand{\doi}[1]{doi: #1}\else
  \providecommand{\doi}{doi: \begingroup \urlstyle{rm}\Url}\fi

\bibitem[Agrawal et~al.(2022)Agrawal, Zhou, Lewis, Zettlemoyer, and
  Ghazvininejad]{agrawal2022incontext}
Sweta Agrawal, Chunting Zhou, Mike Lewis, Luke Zettlemoyer, and Marjan
  Ghazvininejad.
\newblock In-context examples selection for machine translation, 2022.

\bibitem[Bills et~al.(2023)Bills, Cammarata, Mossing, Tillman, Gao, Goh,
  Sutskever, Leike, Wu, and Saunders]{bills2023language}
Steven Bills, Nick Cammarata, Dan Mossing, Henk Tillman, Leo Gao, Gabriel Goh,
  Ilya Sutskever, Jan Leike, Jeff Wu, and William Saunders.
\newblock Language models can explain neurons in language models.
\newblock \emph{URL https://openaipublic. blob. core. windows.
  net/neuron-explainer/paper/index. html.(Date accessed: 14.05. 2023)}, 2023.

\bibitem[Briakou et~al.(2023)Briakou, Cherry, and Foster]{briakou2023searching}
Eleftheria Briakou, Colin Cherry, and George Foster.
\newblock Searching for needles in a haystack: On the role of incidental
  bilingualism in palm's translation capability.
\newblock \emph{arXiv preprint arXiv:2305.10266}, 2023.

\bibitem[Brown et~al.(2020)Brown, Mann, Ryder, Subbiah, Kaplan, Dhariwal,
  Neelakantan, Shyam, Sastry, Askell, Agarwal, Herbert{-}Voss, Krueger,
  Henighan, Child, Ramesh, Ziegler, Wu, Winter, Hesse, Chen, Sigler, Litwin,
  Gray, Chess, Clark, Berner, McCandlish, Radford, Sutskever, and
  Amodei]{DBLP:journals/corr/abs-2005-14165}
Tom~B. Brown, Benjamin Mann, Nick Ryder, Melanie Subbiah, Jared Kaplan,
  Prafulla Dhariwal, Arvind Neelakantan, Pranav Shyam, Girish Sastry, Amanda
  Askell, Sandhini Agarwal, Ariel Herbert{-}Voss, Gretchen Krueger, Tom
  Henighan, Rewon Child, Aditya Ramesh, Daniel~M. Ziegler, Jeffrey Wu, Clemens
  Winter, Christopher Hesse, Mark Chen, Eric Sigler, Mateusz Litwin, Scott
  Gray, Benjamin Chess, Jack Clark, Christopher Berner, Sam McCandlish, Alec
  Radford, Ilya Sutskever, and Dario Amodei.
\newblock Language models are few-shot learners.
\newblock \emph{CoRR}, abs/2005.14165, 2020.
\newblock URL \url{https://arxiv.org/abs/2005.14165}.

\bibitem[Bubeck et~al.(2023)Bubeck, Chandrasekaran, Eldan, Gehrke, Horvitz,
  Kamar, Lee, Lee, Li, Lundberg, et~al.]{bubeck2023sparks}
S{\'e}bastien Bubeck, Varun Chandrasekaran, Ronen Eldan, Johannes Gehrke, Eric
  Horvitz, Ece Kamar, Peter Lee, Yin~Tat Lee, Yuanzhi Li, Scott Lundberg,
  et~al.
\newblock Sparks of artificial general intelligence: Early experiments with
  gpt-4.
\newblock \emph{arXiv preprint arXiv:2303.12712}, 2023.

\bibitem[Chen et~al.(2023{\natexlab{a}})Chen, Borgeaud, Irving, Lespiau, Sifre,
  and Jumper]{chen2023accelerating}
Charlie Chen, Sebastian Borgeaud, Geoffrey Irving, Jean-Baptiste Lespiau,
  Laurent Sifre, and John Jumper.
\newblock Accelerating large language model decoding with speculative sampling.
\newblock \emph{arXiv preprint arXiv:2302.01318}, 2023{\natexlab{a}}.

\bibitem[Chen et~al.(2023{\natexlab{b}})Chen, Guo, Haddow, and
  Heafield]{chen2023iterative}
Pinzhen Chen, Zhicheng Guo, Barry Haddow, and Kenneth Heafield.
\newblock Iterative translation refinement with large language models,
  2023{\natexlab{b}}.

\bibitem[Chen et~al.(2017)Chen, Liu, Cheng, and Li]{chen2017teacher}
Yun Chen, Yang Liu, Yong Cheng, and Victor~OK Li.
\newblock A teacher-student framework for zero-resource neural machine
  translation.
\newblock \emph{arXiv preprint arXiv:1705.00753}, 2017.

\bibitem[Cheng et~al.(2022)Cheng, Gao, Liu, Zhao, and Yan]{cheng2022neural}
Xin Cheng, Shen Gao, Lemao Liu, Dongyan Zhao, and Rui Yan.
\newblock Neural machine translation with contrastive translation memories.
\newblock \emph{arXiv preprint arXiv:2212.03140}, 2022.

\bibitem[Cheng et~al.(2023{\natexlab{a}})Cheng, Lin, Chen, Zhao, and
  Yan]{cheng-etal-2023-decouple}
Xin Cheng, Yankai Lin, Xiuying Chen, Dongyan Zhao, and Rui Yan.
\newblock Decouple knowledge from paramters for plug-and-play language
  modeling.
\newblock In \emph{Findings of the Association for Computational Linguistics:
  ACL 2023}, pp.\  14288--14308, Toronto, Canada, July 2023{\natexlab{a}}.
  Association for Computational Linguistics.
\newblock \doi{10.18653/v1/2023.findings-acl.901}.
\newblock URL \url{https://aclanthology.org/2023.findings-acl.901}.

\bibitem[Cheng et~al.(2023{\natexlab{b}})Cheng, Luo, Chen, Liu, Zhao, and
  Yan]{cheng2023lift}
Xin Cheng, Di~Luo, Xiuying Chen, Lemao Liu, Dongyan Zhao, and Rui Yan.
\newblock Lift yourself up: Retrieval-augmented text generation with self
  memory, 2023{\natexlab{b}}.

\bibitem[Dettmers et~al.(2023)Dettmers, Pagnoni, Holtzman, and
  Zettlemoyer]{dettmers2023qlora}
Tim Dettmers, Artidoro Pagnoni, Ari Holtzman, and Luke Zettlemoyer.
\newblock Qlora: Efficient finetuning of quantized llms.
\newblock \emph{arXiv preprint arXiv:2305.14314}, 2023.

\bibitem[Fan et~al.(2021)Fan, Bhosale, Schwenk, Ma, El{-}Kishky, Goyal, Baines,
  Celebi, Wenzek, Chaudhary, Goyal, Birch, Liptchinsky, Edunov, Auli, and
  Joulin]{DBLP:journals/jmlr/FanBSMEGBCWCGBL21}
Angela Fan, Shruti Bhosale, Holger Schwenk, Zhiyi Ma, Ahmed El{-}Kishky,
  Siddharth Goyal, Mandeep Baines, Onur Celebi, Guillaume Wenzek, Vishrav
  Chaudhary, Naman Goyal, Tom Birch, Vitaliy Liptchinsky, Sergey Edunov,
  Michael Auli, and Armand Joulin.
\newblock Beyond english-centric multilingual machine translation.
\newblock \emph{J. Mach. Learn. Res.}, 22:\penalty0 107:1--107:48, 2021.
\newblock URL \url{http://jmlr.org/papers/v22/20-1307.html}.

\bibitem[Freitag et~al.(2022)Freitag, Rei, Mathur, Lo, Stewart, Avramidis,
  Kocmi, Foster, Lavie, and Martins]{DBLP:conf/wmt/FreitagRMLSAKFLM22}
Markus Freitag, Ricardo Rei, Nitika Mathur, Chi{-}kiu Lo, Craig Stewart,
  Eleftherios Avramidis, Tom Kocmi, George~F. Foster, Alon Lavie, and
  Andr{\'{e}} F.~T. Martins.
\newblock Results of {WMT22} metrics shared task: Stop using {BLEU} - neural
  metrics are better and more robust.
\newblock In Philipp Koehn, Lo{\"{\i}}c Barrault, Ondrej Bojar, Fethi Bougares,
  Rajen Chatterjee, Marta~R. Costa{-}juss{\`{a}}, Christian Federmann, Mark
  Fishel, Alexander Fraser, Markus Freitag, Yvette Graham, Roman Grundkiewicz,
  Paco Guzman, Barry Haddow, Matthias Huck, Antonio Jimeno{-}Yepes, Tom Kocmi,
  Andr{\'{e}} Martins, Makoto Morishita, Christof Monz, Masaaki Nagata,
  Toshiaki Nakazawa, Matteo Negri, Aur{\'{e}}lie N{\'{e}}v{\'{e}}ol, Mariana
  Neves, Martin Popel, Marco Turchi, and Marcos Zampieri (eds.),
  \emph{Proceedings of the Seventh Conference on Machine Translation, {WMT}
  2022, Abu Dhabi, United Arab Emirates (Hybrid), December 7-8, 2022}, pp.\
  46--68. Association for Computational Linguistics, 2022.
\newblock URL \url{https://aclanthology.org/2022.wmt-1.2}.

\bibitem[Garcia et~al.(2023)Garcia, Bansal, Cherry, Foster, Krikun, Feng,
  Johnson, and Firat]{DBLP:journals/corr/abs-2302-01398}
Xavier Garcia, Yamini Bansal, Colin Cherry, George~F. Foster, Maxim Krikun,
  Fangxiaoyu Feng, Melvin Johnson, and Orhan Firat.
\newblock The unreasonable effectiveness of few-shot learning for machine
  translation.
\newblock \emph{CoRR}, abs/2302.01398, 2023.
\newblock \doi{10.48550/arXiv.2302.01398}.
\newblock URL \url{https://doi.org/10.48550/arXiv.2302.01398}.

\bibitem[Ge et~al.(2023)Ge, Hu, Wang, Chen, and Wei]{ge2023context}
Tao Ge, Jing Hu, Xun Wang, Si-Qing Chen, and Furu Wei.
\newblock In-context autoencoder for context compression in a large language
  model.
\newblock \emph{arXiv preprint arXiv:2307.06945}, 2023.

\bibitem[Guerreiro et~al.(2023)Guerreiro, Alves, Waldendorf, Haddow, Birch,
  Colombo, and Martins]{guerreiro2023hallucinations}
Nuno~M. Guerreiro, Duarte Alves, Jonas Waldendorf, Barry Haddow, Alexandra
  Birch, Pierre Colombo, and André F.~T. Martins.
\newblock Hallucinations in large multilingual translation models, 2023.

\bibitem[Hendy et~al.(2023)Hendy, Abdelrehim, Sharaf, Raunak, Gabr, Matsushita,
  Kim, Afify, and Awadalla]{DBLP:journals/corr/abs-2302-09210}
Amr Hendy, Mohamed Abdelrehim, Amr Sharaf, Vikas Raunak, Mohamed Gabr, Hitokazu
  Matsushita, Young~Jin Kim, Mohamed Afify, and Hany~Hassan Awadalla.
\newblock How good are {GPT} models at machine translation? {A} comprehensive
  evaluation.
\newblock \emph{CoRR}, abs/2302.09210, 2023.
\newblock \doi{10.48550/arXiv.2302.09210}.
\newblock URL \url{https://doi.org/10.48550/arXiv.2302.09210}.

\bibitem[Hu et~al.(2021)Hu, Shen, Wallis, Allen-Zhu, Li, Wang, Wang, and
  Chen]{hu2021lora}
Edward~J. Hu, Yelong Shen, Phillip Wallis, Zeyuan Allen-Zhu, Yuanzhi Li, Shean
  Wang, Lu~Wang, and Weizhu Chen.
\newblock Lora: Low-rank adaptation of large language models, 2021.

\bibitem[Jiao et~al.(2023)Jiao, Wang, Huang, Wang, and Tu]{jiao2023chatgpt}
Wenxiang Jiao, Wenxuan Wang, Jen-tse Huang, Xing Wang, and Zhaopeng Tu.
\newblock Is chatgpt a good translator? a preliminary study.
\newblock \emph{arXiv preprint arXiv:2301.08745}, 2023.

\bibitem[Kim et~al.(2019)Kim, Petrov, Petrushkov, Khadivi, and
  Ney]{kim2019pivot}
Yunsu Kim, Petre Petrov, Pavel Petrushkov, Shahram Khadivi, and Hermann Ney.
\newblock Pivot-based transfer learning for neural machine translation between
  non-english languages.
\newblock \emph{arXiv preprint arXiv:1909.09524}, 2019.

\bibitem[Lin et~al.(2022)Lin, Mihaylov, Artetxe, Wang, Chen, Simig, Ott, Goyal,
  Bhosale, Du, Pasunuru, Shleifer, Koura, Chaudhary, O'Horo, Wang, Zettlemoyer,
  Kozareva, Diab, Stoyanov, and Li]{DBLP:conf/emnlp/LinMAWCSOGBDPSK22}
Xi~Victoria Lin, Todor Mihaylov, Mikel Artetxe, Tianlu Wang, Shuohui Chen,
  Daniel Simig, Myle Ott, Naman Goyal, Shruti Bhosale, Jingfei Du, Ramakanth
  Pasunuru, Sam Shleifer, Punit~Singh Koura, Vishrav Chaudhary, Brian O'Horo,
  Jeff Wang, Luke Zettlemoyer, Zornitsa Kozareva, Mona~T. Diab, Veselin
  Stoyanov, and Xian Li.
\newblock Few-shot learning with multilingual generative language models.
\newblock In Yoav Goldberg, Zornitsa Kozareva, and Yue Zhang (eds.),
  \emph{Proceedings of the 2022 Conference on Empirical Methods in Natural
  Language Processing, {EMNLP} 2022, Abu Dhabi, United Arab Emirates, December
  7-11, 2022}, pp.\  9019--9052. Association for Computational Linguistics,
  2022.
\newblock URL \url{https://aclanthology.org/2022.emnlp-main.616}.

\bibitem[Lin et~al.(2023)Lin, Tan, Lin, Zheng, Pi, Zhang, Diao, Wang, Zhao,
  Yao, and Zhang]{lin2023speciality}
Yong Lin, Lu~Tan, Hangyu Lin, Zeming Zheng, Renjie Pi, Jipeng Zhang, Shizhe
  Diao, Haoxiang Wang, Han Zhao, Yuan Yao, and Tong Zhang.
\newblock Speciality vs generality: An empirical study on catastrophic
  forgetting in fine-tuning foundation models, 2023.

\bibitem[Luo et~al.(2023)Luo, Xu, Zhao, Geng, Tao, Ma, Lin, and
  Jiang]{luo2023augmented}
Ziyang Luo, Can Xu, Pu~Zhao, Xiubo Geng, Chongyang Tao, Jing Ma, Qingwei Lin,
  and Daxin Jiang.
\newblock Augmented large language models with parametric knowledge guiding,
  2023.

\bibitem[{NLLB Team} et~al.(2022){NLLB Team}, Costa-jussà, Cross, Çelebi,
  Elbayad, Heafield, Heffernan, Kalbassi, Lam, Licht, Maillard, Sun, Wang,
  Wenzek, Youngblood, Akula, Barrault, Mejia-Gonzalez, Hansanti, Hoffman,
  Jarrett, Sadagopan, Rowe, Spruit, Tran, Andrews, Ayan, Bhosale, Edunov, Fan,
  Gao, Goswami, Guzmán, Koehn, Mourachko, Ropers, Saleem, Schwenk, and
  Wang]{nllb2022}
{NLLB Team}, Marta~R. Costa-jussà, James Cross, Onur Çelebi, Maha Elbayad,
  Kenneth Heafield, Kevin Heffernan, Elahe Kalbassi, Janice Lam, Daniel Licht,
  Jean Maillard, Anna Sun, Skyler Wang, Guillaume Wenzek, Al~Youngblood, Bapi
  Akula, Loic Barrault, Gabriel Mejia-Gonzalez, Prangthip Hansanti, John
  Hoffman, Semarley Jarrett, Kaushik~Ram Sadagopan, Dirk Rowe, Shannon Spruit,
  Chau Tran, Pierre Andrews, Necip~Fazil Ayan, Shruti Bhosale, Sergey Edunov,
  Angela Fan, Cynthia Gao, Vedanuj Goswami, Francisco Guzmán, Philipp Koehn,
  Alexandre Mourachko, Christophe Ropers, Safiyyah Saleem, Holger Schwenk, and
  Jeff Wang.
\newblock No language left behind: Scaling human-centered machine translation.
\newblock 2022.

\bibitem[OpenAI(2023)]{OpenAI2023GPT4TR}
OpenAI.
\newblock Gpt-4 technical report.
\newblock \emph{ArXiv}, abs/2303.08774, 2023.
\newblock URL \url{https://api.semanticscholar.org/CorpusID:257532815}.

\bibitem[Peng et~al.(2023)Peng, Alcaide, Anthony, Albalak, Arcadinho, Cao,
  Cheng, Chung, Grella, GV, et~al.]{peng2023rwkv}
Bo~Peng, Eric Alcaide, Quentin Anthony, Alon Albalak, Samuel Arcadinho, Huanqi
  Cao, Xin Cheng, Michael Chung, Matteo Grella, Kranthi~Kiran GV, et~al.
\newblock Rwkv: Reinventing rnns for the transformer era.
\newblock \emph{arXiv preprint arXiv:2305.13048}, 2023.

\bibitem[Popovic(2017)]{DBLP:conf/wmt/Popovic17}
Maja Popovic.
\newblock chrf++: words helping character n-grams.
\newblock In Ondrej Bojar, Christian Buck, Rajen Chatterjee, Christian
  Federmann, Yvette Graham, Barry Haddow, Matthias Huck, Antonio
  Jimeno{-}Yepes, Philipp Koehn, and Julia Kreutzer (eds.), \emph{Proceedings
  of the Second Conference on Machine Translation, {WMT} 2017, Copenhagen,
  Denmark, September 7-8, 2017}, pp.\  612--618. Association for Computational
  Linguistics, 2017.
\newblock \doi{10.18653/v1/w17-4770}.
\newblock URL \url{https://doi.org/10.18653/v1/w17-4770}.

\bibitem[Press et~al.(2022)Press, Zhang, Min, Schmidt, Smith, and
  Lewis]{press2022measuring}
Ofir Press, Muru Zhang, Sewon Min, Ludwig Schmidt, Noah~A Smith, and Mike
  Lewis.
\newblock Measuring and narrowing the compositionality gap in language models.
\newblock \emph{arXiv preprint arXiv:2210.03350}, 2022.

\bibitem[Radford et~al.(2019)Radford, Wu, Child, Luan, Amodei, and
  Sutskever]{Radford2019LanguageMA}
Alec Radford, Jeff Wu, Rewon Child, David Luan, Dario Amodei, and Ilya
  Sutskever.
\newblock Language models are unsupervised multitask learners.
\newblock 2019.
\newblock URL \url{https://api.semanticscholar.org/CorpusID:160025533}.

\bibitem[Raunak et~al.(2022)Raunak, Post, and Menezes]{raunak-etal-2022-salted}
Vikas Raunak, Matt Post, and Arul Menezes.
\newblock {SALTED}: A framework for {SA}lient long-tail translation error
  detection.
\newblock In \emph{Findings of the Association for Computational Linguistics:
  EMNLP 2022}, pp.\  5163--5179, Abu Dhabi, United Arab Emirates, December
  2022. Association for Computational Linguistics.
\newblock \doi{10.18653/v1/2022.findings-emnlp.379}.
\newblock URL \url{https://aclanthology.org/2022.findings-emnlp.379}.

\bibitem[Raunak et~al.(2023{\natexlab{a}})Raunak, Menezes, Post, and
  Awadalla]{raunak2023gpts}
Vikas Raunak, Arul Menezes, Matt Post, and Hany~Hassan Awadalla.
\newblock Do gpts produce less literal translations?, 2023{\natexlab{a}}.

\bibitem[Raunak et~al.(2023{\natexlab{b}})Raunak, Sharaf, Awadallah, and
  Menezes]{raunak2023leveraging}
Vikas Raunak, Amr Sharaf, Hany~Hassan Awadallah, and Arul Menezes.
\newblock Leveraging gpt-4 for automatic translation post-editing,
  2023{\natexlab{b}}.

\bibitem[Rei et~al.(2020)Rei, Stewart, Farinha, and
  Lavie]{DBLP:conf/emnlp/ReiSFL20}
Ricardo Rei, Craig Stewart, Ana~C. Farinha, and Alon Lavie.
\newblock {COMET:} {A} neural framework for {MT} evaluation.
\newblock In Bonnie Webber, Trevor Cohn, Yulan He, and Yang Liu (eds.),
  \emph{Proceedings of the 2020 Conference on Empirical Methods in Natural
  Language Processing, {EMNLP} 2020, Online, November 16-20, 2020}, pp.\
  2685--2702. Association for Computational Linguistics, 2020.
\newblock \doi{10.18653/v1/2020.emnlp-main.213}.
\newblock URL \url{https://doi.org/10.18653/v1/2020.emnlp-main.213}.

\bibitem[Rei et~al.(2022{\natexlab{a}})Rei, de~Souza, Alves, Zerva, Farinha,
  Glushkova, Lavie, Coheur, and Martins]{DBLP:conf/wmt/ReiSAZFGLCM22}
Ricardo Rei, Jos{\'{e}} G.~C. de~Souza, Duarte~M. Alves, Chrysoula Zerva,
  Ana~C. Farinha, Taisiya Glushkova, Alon Lavie, Lu{\'{\i}}sa Coheur, and
  Andr{\'{e}} F.~T. Martins.
\newblock {COMET-22:} unbabel-ist 2022 submission for the metrics shared task.
\newblock In Philipp Koehn, Lo{\"{\i}}c Barrault, Ondrej Bojar, Fethi Bougares,
  Rajen Chatterjee, Marta~R. Costa{-}juss{\`{a}}, Christian Federmann, Mark
  Fishel, Alexander Fraser, Markus Freitag, Yvette Graham, Roman Grundkiewicz,
  Paco Guzman, Barry Haddow, Matthias Huck, Antonio Jimeno{-}Yepes, Tom Kocmi,
  Andr{\'{e}} Martins, Makoto Morishita, Christof Monz, Masaaki Nagata,
  Toshiaki Nakazawa, Matteo Negri, Aur{\'{e}}lie N{\'{e}}v{\'{e}}ol, Mariana
  Neves, Martin Popel, Marco Turchi, and Marcos Zampieri (eds.),
  \emph{Proceedings of the Seventh Conference on Machine Translation, {WMT}
  2022, Abu Dhabi, United Arab Emirates (Hybrid), December 7-8, 2022}, pp.\
  578--585. Association for Computational Linguistics, 2022{\natexlab{a}}.
\newblock URL \url{https://aclanthology.org/2022.wmt-1.52}.

\bibitem[Rei et~al.(2022{\natexlab{b}})Rei, Treviso, Guerreiro, Zerva, Farinha,
  Maroti, C.~de Souza, Glushkova, Alves, Coheur, Lavie, and
  Martins]{rei-etal-2022-cometkiwi}
Ricardo Rei, Marcos Treviso, Nuno~M. Guerreiro, Chrysoula Zerva, Ana~C Farinha,
  Christine Maroti, Jos{\'e}~G. C.~de Souza, Taisiya Glushkova, Duarte Alves,
  Luisa Coheur, Alon Lavie, and Andr{\'e} F.~T. Martins.
\newblock {C}omet{K}iwi: {IST}-unbabel 2022 submission for the quality
  estimation shared task.
\newblock In \emph{Proceedings of the Seventh Conference on Machine Translation
  (WMT)}, pp.\  634--645, Abu Dhabi, United Arab Emirates (Hybrid), December
  2022{\natexlab{b}}. Association for Computational Linguistics.
\newblock URL \url{https://aclanthology.org/2022.wmt-1.60}.

\bibitem[Scao et~al.(2022)Scao, Fan, Akiki, Pavlick, Ili{\'c}, Hesslow,
  Castagn{\'e}, Luccioni, Yvon, Gall{\'e}, et~al.]{scao2022bloom}
Teven~Le Scao, Angela Fan, Christopher Akiki, Ellie Pavlick, Suzana Ili{\'c},
  Daniel Hesslow, Roman Castagn{\'e}, Alexandra~Sasha Luccioni, Fran{\c{c}}ois
  Yvon, Matthias Gall{\'e}, et~al.
\newblock Bloom: A 176b-parameter open-access multilingual language model.
\newblock \emph{arXiv preprint arXiv:2211.05100}, 2022.

\bibitem[Schioppa et~al.(2021)Schioppa, Vilar, Sokolov, and
  Filippova]{schioppa-etal-2021-controlling}
Andrea Schioppa, David Vilar, Artem Sokolov, and Katja Filippova.
\newblock Controlling machine translation for multiple attributes with additive
  interventions.
\newblock In \emph{Proceedings of the 2021 Conference on Empirical Methods in
  Natural Language Processing}, pp.\  6676--6696, Online and Punta Cana,
  Dominican Republic, November 2021. Association for Computational Linguistics.
\newblock \doi{10.18653/v1/2021.emnlp-main.535}.
\newblock URL \url{https://aclanthology.org/2021.emnlp-main.535}.

\bibitem[Sellam et~al.(2020)Sellam, Das, and Parikh]{DBLP:conf/acl/SellamDP20}
Thibault Sellam, Dipanjan Das, and Ankur~P. Parikh.
\newblock {BLEURT:} learning robust metrics for text generation.
\newblock In Dan Jurafsky, Joyce Chai, Natalie Schluter, and Joel~R. Tetreault
  (eds.), \emph{Proceedings of the 58th Annual Meeting of the Association for
  Computational Linguistics, {ACL} 2020, Online, July 5-10, 2020}, pp.\
  7881--7892. Association for Computational Linguistics, 2020.
\newblock \doi{10.18653/v1/2020.acl-main.704}.
\newblock URL \url{https://doi.org/10.18653/v1/2020.acl-main.704}.

\bibitem[Sutskever et~al.(2014)Sutskever, Vinyals, and
  Le]{sutskever2014sequence}
Ilya Sutskever, Oriol Vinyals, and Quoc~V Le.
\newblock Sequence to sequence learning with neural networks.
\newblock \emph{Advances in neural information processing systems}, 27, 2014.

\bibitem[Touvron et~al.(2023)Touvron, Lavril, Izacard, Martinet, Lachaux,
  Lacroix, Rozière, Goyal, Hambro, Azhar, Rodriguez, Joulin, Grave, and
  Lample]{touvron2023llama}
Hugo Touvron, Thibaut Lavril, Gautier Izacard, Xavier Martinet, Marie-Anne
  Lachaux, Timothée Lacroix, Baptiste Rozière, Naman Goyal, Eric Hambro,
  Faisal Azhar, Aurelien Rodriguez, Armand Joulin, Edouard Grave, and Guillaume
  Lample.
\newblock Llama: Open and efficient foundation language models, 2023.

\bibitem[Vaswani et~al.(2017)Vaswani, Shazeer, Parmar, Uszkoreit, Jones, Gomez,
  Kaiser, and Polosukhin]{vaswani2017attention}
Ashish Vaswani, Noam Shazeer, Niki Parmar, Jakob Uszkoreit, Llion Jones,
  Aidan~N Gomez, {\L}ukasz Kaiser, and Illia Polosukhin.
\newblock Attention is all you need.
\newblock \emph{Advances in neural information processing systems}, 30, 2017.

\bibitem[Vilar et~al.(2023)Vilar, Freitag, Cherry, Luo, Ratnakar, and
  Foster]{vilar2023prompting}
David Vilar, Markus Freitag, Colin Cherry, Jiaming Luo, Viresh Ratnakar, and
  George Foster.
\newblock Prompting palm for translation: Assessing strategies and performance,
  2023.

\bibitem[Wei et~al.(2023)Wei, Wei, Tay, Tran, Webson, Lu, Chen, Liu, Huang,
  Zhou, et~al.]{wei2023larger}
Jerry Wei, Jason Wei, Yi~Tay, Dustin Tran, Albert Webson, Yifeng Lu, Xinyun
  Chen, Hanxiao Liu, Da~Huang, Denny Zhou, et~al.
\newblock Larger language models do in-context learning differently.
\newblock \emph{arXiv preprint arXiv:2303.03846}, 2023.

\bibitem[Xia et~al.(2022)Xia, Ge, Chen, Wei, and Sui]{xia2022speculative}
Heming Xia, Tao Ge, Si-Qing Chen, Furu Wei, and Zhifang Sui.
\newblock Speculative decoding: Lossless speedup of autoregressive translation.
\newblock 2022.
\newblock URL \url{https://openreview.net/pdf?id=H-VlwsYvVi}.

\bibitem[Xia et~al.(2017)Xia, Tian, Wu, Lin, Qin, Yu, and
  Liu]{xia2017deliberation}
Yingce Xia, Fei Tian, Lijun Wu, Jianxin Lin, Tao Qin, Nenghai Yu, and Tie-Yan
  Liu.
\newblock Deliberation networks: Sequence generation beyond one-pass decoding.
\newblock \emph{Advances in neural information processing systems}, 30, 2017.

\bibitem[Xu et~al.(2023)Xu, Xu, Wang, Liu, Zhu, and McAuley]{xu2023small}
Canwen Xu, Yichong Xu, Shuohang Wang, Yang Liu, Chenguang Zhu, and Julian
  McAuley.
\newblock Small models are valuable plug-ins for large language models.
\newblock \emph{arXiv preprint arXiv:2305.08848}, 2023.

\bibitem[Yang et~al.(2023)Yang, Ge, Wang, Jiao, Jiang, Yang, Majumder, and
  Wei]{yang2023inference}
Nan Yang, Tao Ge, Liang Wang, Binxing Jiao, Daxin Jiang, Linjun Yang, Rangan
  Majumder, and Furu Wei.
\newblock Inference with reference: Lossless acceleration of large language
  models, 2023.

\bibitem[Yao et~al.(2023)Yao, Jiang, Yang, and Hu]{yao2023empowering}
Binwei Yao, Ming Jiang, Diyi Yang, and Junjie Hu.
\newblock Empowering llm-based machine translation with cultural awareness.
\newblock \emph{arXiv preprint arXiv:2305.14328}, 2023.

\bibitem[Yong et~al.(2023)Yong, Schoelkopf, Muennighoff, Aji, Adelani,
  Almubarak, Bari, Sutawika, Kasai, Baruwa, Winata, Biderman, Raff, Radev, and
  Nikoulina]{yong2023bloom1}
Zheng-Xin Yong, Hailey Schoelkopf, Niklas Muennighoff, Alham~Fikri Aji,
  David~Ifeoluwa Adelani, Khalid Almubarak, M~Saiful Bari, Lintang Sutawika,
  Jungo Kasai, Ahmed Baruwa, Genta~Indra Winata, Stella Biderman, Edward Raff,
  Dragomir Radev, and Vassilina Nikoulina.
\newblock Bloom+1: Adding language support to bloom for zero-shot prompting,
  2023.

\bibitem[Zeng et~al.(2023)Zeng, Meng, Yin, and Zhou]{zeng2023tim}
Jiali Zeng, Fandong Meng, Yongjing Yin, and Jie Zhou.
\newblock Tim: Teaching large language models to translate with comparison.
\newblock \emph{arXiv preprint arXiv:2307.04408}, 2023.

\bibitem[Zhu et~al.(2023)Zhu, Liu, Dong, Xu, Huang, Kong, Chen, and
  Li]{zhu2023multilingual}
Wenhao Zhu, Hongyi Liu, Qingxiu Dong, Jingjing Xu, Shujian Huang, Lingpeng
  Kong, Jiajun Chen, and Lei Li.
\newblock Multilingual machine translation with large language models:
  Empirical results and analysis, 2023.

\end{thebibliography}
